\documentclass[runningheads]{llncs}
\usepackage{graphicx}
\usepackage{todonotes}
\usepackage{url}
\usepackage{cite}
\usepackage{bbm}
\usepackage{amsmath}
\usepackage{amssymb}
\usepackage{bbold}
\usepackage{multirow}
\usepackage{algorithm,algorithmic}
\usepackage{comment}
\usepackage{graphicx}
\usepackage{bbm}
\usepackage{bm}
\usepackage{dsfont}
\usepackage{mathtools}
\usepackage{adjustbox}
\usepackage{hyperref}

\begin{document}
\title{HSIC-InfoGAN: Learning Unsupervised Disentangled Representations by Maximising Approximated Mutual Information}
\titlerunning{HSIC-InfoGAN}

\author{Xiao Liu\inst{1,4}, Spyridon Thermos\inst{2}, Pedro Sanchez\inst{1,4}, Alison Q. O'Neil\inst{1,4} \and Sotirios A. Tsaftaris\inst{1,3,4}} 
\authorrunning{X. Liu et al.}
\institute{School of Engineering, University of Edinburgh, Edinburgh EH9 3FB, UK 
\and
AC Codewheel Ltd
\and
The Alan Turing Institute, London, UK \and
Canon Medical Research Europe Ltd., Edinburgh, UK\\
\email{Xiao.Liu@ed.ac.uk}}


%
\maketitle              
\sloppy
\setcounter{footnote}{0} 
\begin{abstract}
Learning disentangled representations requires either supervision or the introduction of specific model designs and learning constraints as biases. InfoGAN is a popular disentanglement framework that learns unsupervised disentangled representations by maximising the mutual information between latent representations and their corresponding generated images. Maximisation of mutual information is achieved by introducing an auxiliary network and training with a latent regression loss. In this short exploratory paper, we study the use of the Hilbert-Schmidt Independence Criterion (HSIC) to approximate mutual information between latent representation and image, termed HSIC-InfoGAN. Directly optimising the HSIC loss avoids the need for an additional auxiliary network. We qualitatively compare the level of disentanglement in each model, suggest a strategy to tune the hyperparameters of HSIC-InfoGAN, and discuss the potential of HSIC-InfoGAN for medical applications.

\end{abstract}

\section{Introduction}
Recently, machine learning (ML) and deep learning (DL) have achieved significant success in many computer science areas, for instance vision and natural language processing. \cite{lecun2015deep}. However, traditional fully supervised approaches cannot always be applied in specific domains such as medical imaging analysis, as the available annotations are limited due to the labeling process being tedious and costly. Thus, significant effort has been placed on alternative training methods such as unsupervised and semi-supervised learning. In particular, recent works \cite{kingma2013auto, liu2020disentangled, chartsias2019disentangled} show that the typical unsupervised approach of disentangled representation learning without labeled data significantly boosts the performance of ML/DL models. 

The widely agreed definition of a disentangled representation is one in which ``\textit{single latent units are sensitive to changes in single generative factors, while being relatively invariant to changes in other factors}'' \cite{bengio2013representation}. This definition is based on an implicit assumption that there is a generation process in the real world that translates independent generative factors to images. Hence, the overall goal in disentangled representation learning is to discover this generation process and the constituent generative factors from images. However, a comprehensive study \cite{locatello2019challenging} recently showed that it is impossible to learn a disentangled representation in an unsupervised setting, and all previous methods use various inductive biases or assumptions on either model design or learning process. Then, model performance largely depends on the introduced inductive biases that are tailored for specific tasks. In other words, different tasks require domain-specific expert knowledge to devise suitable inductive biases and assumptions.

As an unsupervised disentanglement method, InfoGAN \cite{chen2016infogan} considers that the latent representations consist of categorical (we assume that we know the number of classes) and continuous latents. With this inductive bias, InfoGAN solves the information-regularised minimax game by jointly training a generator, a discriminator and an auxiliary network. In particular, the auxiliary network takes the generated image as input and is trained to correctly predict the corresponding latent representations of this image i.e.\ latent regression \cite{zhu2017toward}. This forces the generated image to be highly dependent on the latent representations i.e.\ maximising the mutual information. Despite the cost of introducing an additional network/module, this strategy has been used in many disentanglement models such as \cite{zhu2017toward, huang2018multimodal, odena2017conditional}. 

In this paper, we examine the question of whether we can approximate mutual information between the latent representations and the generated image without the need for such an auxiliary network. We consider the Hilbert-Schmidt Independence Criterion (HSIC) \cite{ma2020hsic} to approximate the mutual information between the latent representations and the generated image, termed HSIC-InfoGAN. HSIC is a kernel-based independence measurement. By projecting the inputs into kernel space, HSIC allows inputs to have different dimensionality. Directly optimising the HSIC loss removes the need for an auxiliary network, which could reduce the model training time and the memory load for saving model weights. As shown in our experiments, achieves satisfactory levels of disentanglement compared to InfoGAN. We further discuss a strategy to effectively tune the hyperparameters of HSIC-InfoGAN, and discuss its potential impact on medical applications.

\section{Methodology}
\label{sec:method}
\subsection{InfoGAN}
Generative adversarial networks \cite{goodfellow2014generative} train the generator $G$ and discriminator $D$ using a minimax game by optimising the following objective:

\begin{equation}
    \min_{G} \max_{D} V(D, G) = \mathbb{E}_{\mathbf{X}}[\log D(\mathbf{X})] +  \mathbb{E}_{\mathbf{z}}[\log (1 - D(G(\mathbf{z})))],
\end{equation}
where $\mathbf{X}$ denotes an image sample and $\mathbf{z}$ is the noise vector. InfoGAN \cite{chen2016infogan} proposes that the latent space contains the noise $\mathbf{z}$ as well as the disentangled latent code $\mathbf{c}$. To learn the disentangled representations, InfoGAN solves the information-regularised minimax game:
\begin{equation}
    \min_{G} \max_{D} V_{I}(D, G) = V(D, G)-\lambda_{I}I(\mathbf{c}; G(\mathbf{z}, \mathbf{c})),
\end{equation}
where $I(\mathbf{c}; G(\mathbf{z}, \mathbf{c}))$ denotes the mutual information between the latent code $\mathbf{c}$ and the generated image $G(\mathbf{z}, \mathbf{c})$. However, we can only compute the exact and tractable mutual information for discrete variables or for specific problems that we know the probability distributions \cite{belghazi2018mutual}. Due to the difficulty of directly maximising the mutual information term, InfoGAN introduces an auxiliary network $Q$ to derive the lower bound of the mutual information:
\begin{equation}
    I(\mathbf{c}; G(\mathbf{z}, \mathbf{c})) \geq L_I(G, Q) = \mathbb{E}_{\mathbf{X}}[\mathbb{E}_{\mathbf{\mathbf{c}}'}[\log Q(\mathbf{\mathbf{c}}'|\mathbf{X})]].
\end{equation}
Overall, the objective of InfoGAN is defined as:
\begin{equation}
    \min_{G, Q} \max_{D} V_{\textrm{InfoGAN}}(D, G, Q) = V(D, G)-\lambda_{\textrm{InfoGAN}}L_I(G, Q).
\end{equation}
Note that most of the network weights of $Q$ and $D$ can be shared. Separate final (head) layers are used for the $Q$ and $D$ networks in InfoGAN.

\subsection{Hilbert-Schmidt Independence Criterion (HSIC)}
Considering the kernel function $k(,)$, the HSIC loss is defined in \cite{ma2020hsic} as:
\begin{equation}
\label{eq::HSIC}
    \textrm{HSIC}(\mathbf{X}, \mathbf{z}) = (m - 1)^{-2}\textrm{trace}(K_\mathbf{X}HK_\mathbf{z}H),
\end{equation}
where $m$ is the batch size in our case. $K_{\mathbf{X}_{ij}}=k(\mathbf{X}_i, \mathbf{X}_j)$ and $K_{\mathbf{z}_{ij}}=k(\mathbf{z}_i, \mathbf{z}_j)$ are the entries of $K_\mathbf{X} \in R^{m\times m}$ and $K_\mathbf{z} \in R^{m\times m}$. $H$ is the centering matrix $H=I_m - \frac{1}{m}\mathbb{1}_m\mathbb{1}_m^T$. Following \cite{ma2020hsic}, we choose the Gaussian kernel $k(\mathbf{X}_i, \mathbf{X}_j) \sim exp(-\frac{1}{2}||\mathbf{X}_i-\mathbf{X}_j||^2/\sigma^2)$, where $\sigma$ is a hyperparameter. Here, HSIC values are always positive and lower HSIC means higher independence (i.e.\ lower mutual information). Note that in Eq.\ \ref{eq::HSIC}, $\mathbf{X}$ and $\mathbf{z}$ can have different dimensionality i.e.\ they can be a tensor and a vector. We refer the readers to the section ``Relating HSIC to Entropy'' in \cite{ma2020hsic} for a informal discussion about the relationship between HSIC and mutual information. Overall, mutual information is defined in terms of entropy that is related to volume \cite{cover2006elements}, which can be considered as the product of the eigenvalues of the covariance matrix. HSIC is related to Frobenius norm that is a sum of the eigenvalues. 

\subsection{HSIC-InfoGAN}
As discussed in \cite{ma2020hsic}, the mutual information can be approximated with HSIC. We propose to replace the mutual information term in InfoGAN with HSIC as an alternative. Using the HSIC loss we drop the need for an auxiliary network. It could potentially contribute to the stabilisation of InfoGAN training as there is no need for sharing the network weights of discriminator with the auxiliary network. Overall, HSIC-InfoGAN can be represented as:
\begin{equation}
\begin{aligned}
    \min_{G} \max_{D}\ &V_{\textrm{HSIC}}(D, G) = \\ & \mathbb{E}_{\mathbf{X}}[\log D(\mathbf{X})] +  \mathbb{E}_{\mathbf{z}, \mathbf{c}}[\log (1 - D(G(\mathbf{z}, \mathbf{c})))] 
     - \lambda \textrm{HSIC}(\mathbf{X}, \mathbf{c}),
\end{aligned}
\end{equation}
where $\lambda$ is the weight of the HSIC loss. For HSIC-InfoGAN, the tunable hyperparameters are the kernel variance $\sigma$ and loss weight $\lambda$. We will discuss the strategy to tune the two hyperparameters in detail in Section \ref{sec::experiments}.

\section{Experiments}
\label{sec::experiments}
\subsection{Implementation details}
We perform experiments using the MNIST dataset \cite{lecun1998gradient} that contains 60,000 images of 10 digits with image size $28\times 28$. All models are trained using the Adam optimiser \cite{kingma2014adam} with a learning rate of $2\times e^{-4}$ for the discriminator and $1\times e^{-3}$ for the generator. Batch size is 100. We train the models for 100 epochs. Following \cite{chen2016infogan}, we set the dimension of $\mathbf{z}$ to 62 and the dimension of $\mathbf{c}$ to 12, where 10 dimensions of $\mathbf{c}$ represent categorical information (a 10-dimensional one-hot vector) and 2 dimensions of $\mathbf{c}$ represent continuous information (sampled from a uniform distribution $U(0, 1)$). All models are implemented in PyTorch \cite{paszke2019pytorch} and are trained using an NVIDIA 2080 Ti GPU. The code for calculating the HSIC loss can be found in \url{https://github.com/choasma/HSIC-Bottleneck}.

 \subsection{Results}
\begin{figure}[t]
\includegraphics[width=\textwidth]{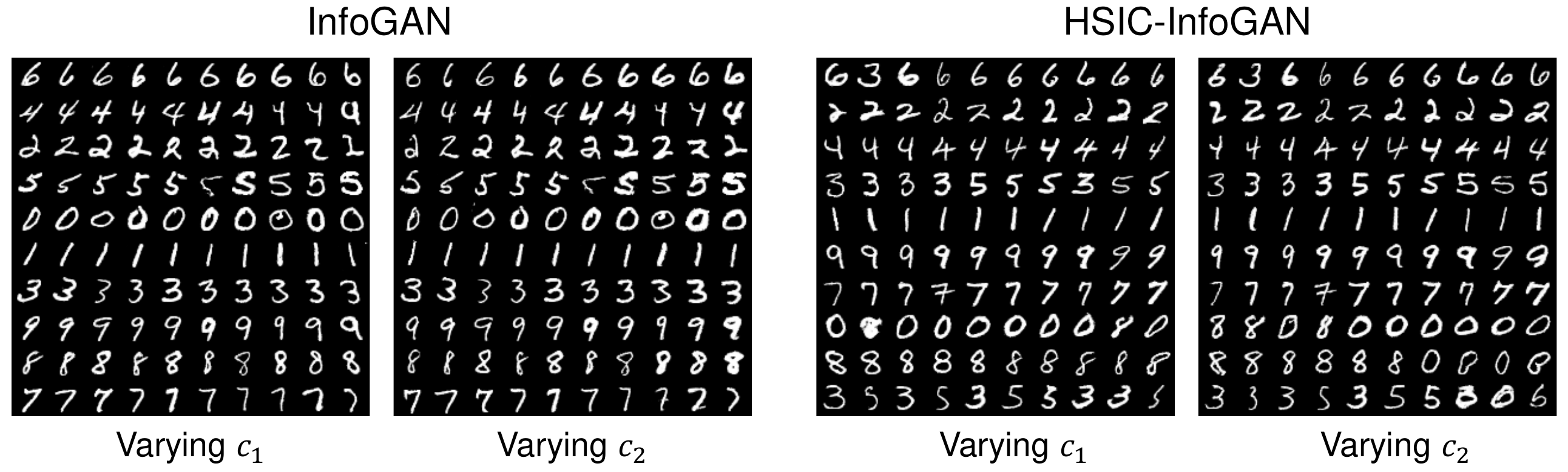}
\centering
\caption{Visual results from InfoGAN and HSIC-InfoGAN.} \label{fig::comparison}
\end{figure}

In Fig. \ref{fig::comparison}, we share qualitative results for InfoGAN and HSIC-InfoGAN. $c_1$ and $c_2$ are the two continuous latent codes. For each row, the categorical latent codes are the same for all the 10 images. For each column,  the continuous latent codes are the same for the 10 images. For each row, we traverse/vary $c_1$ and $c_2$ from $-1$ to $1$. We observe that InfoGAN disentangles nicely the discrete latent codes whilst HSIC-InfoGAN achieves a satisfactory level of disentanglement. When varying $c_2$, HSIC-InfoGAN mixes digit $0$ and $8$ as well as digit $3$ and $5$. Considering $c_1$ mostly captures rotation information and $c_2$ mostly captures thickness information, we observe that HSIC-InfoGAN learns better $c_1$ and similar $c_2$ compared to InfoGAN. Overall, HSIC-InfoGAN achieves satisfactory performance on unsupervised learning of disentangled representations. Considering the benefits of avoiding introducing auxiliary networks, HSIC-InfoGAN offers a good alternative to InfoGAN.

\subsection{Strategy for hyperparameter tuning}
\begin{figure}[t]
\includegraphics[width=\textwidth]{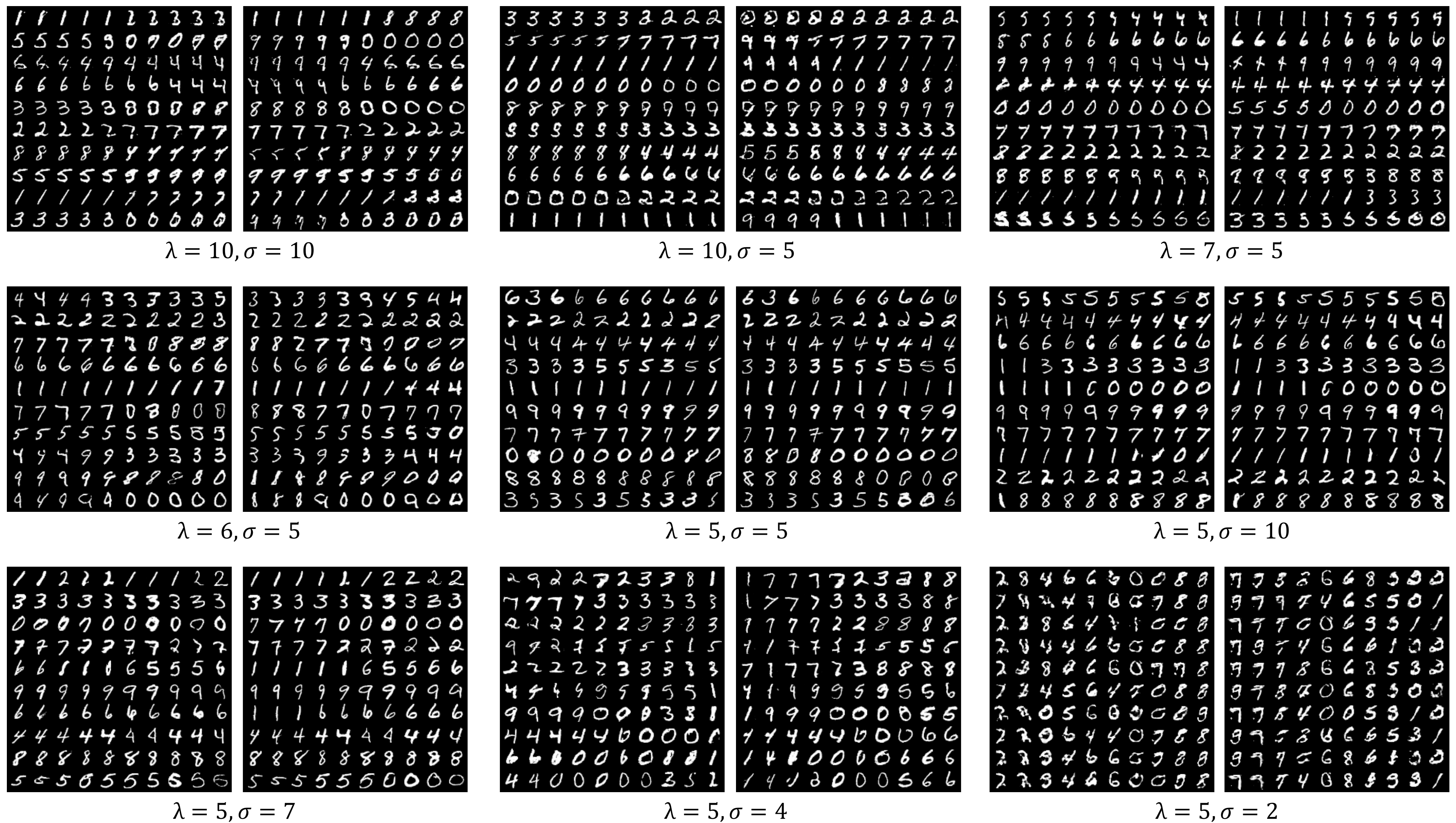}
\centering

\caption{Qualitative results of finding the optimal $\lambda$ and $\sigma$ for HSIC-InfoGAN.} \label{fig::ablation}
\end{figure}

As we discussed in Section \ref{sec:method}, the tunable hyperparameters are the HSIC loss weight $\lambda$ and the kernel variance $\sigma$. We observe that it is important to ensure the generator loss and the HSIC loss have the same order of magnitude. Note that increasing $\lambda$ or decreasing $\sigma$ increase the HSIC loss. As shown in Fig. \ref{fig::ablation}, we show the results of varying $\lambda$ and $\sigma$. Changing $\sigma$ causes more significant changes to the HSIC loss (roughly, we can consider that HSIC $\sim exp(-\frac{1}{\sigma^2})$). Hence, the strategy is to find a good $\sigma$ first (search from $\sigma=2$ to $\sigma=10$ in our case) and then fine tune by finding the optimal $\lambda$.

\section{Discussion}
HSIC-InfoGAN could be widely used in many medical applications. As discussed in \cite{liu2022learning}, many disentanglement methods in the medical domain take advantage of content-style disentanglement \cite{chartsias2019disentangled, liu2021semi, liu2020disentangled, yang2019unsupervised, thermos2021controllable}. In this case, the content and style representations are usually a tensor and a vector \cite{liu2020metrics}. HSIC-InfoGAN can be applied as an alternative to replace the (InfoGAN-style) latent regression losses to save training time and decrease the memory requirements for model weights. In addition, considering other generative models such as normalising flows \cite{papamakarios2021normalizing}, energy-based models \cite{du2021unsupervised} and denoising diffusion models \cite{sanchez2021diffusion}, the HSIC loss (approximating the mutual information) could be used as an unsupervised objective to learn the disentangled latent representations for medical applications. Finally, we envision that HSIC-InfoGAN can be applied in the context of contrastive learning for medical applications \cite{chaitanya2020contrastive}, where one can maximise and minimise the mutual information between features of (different) images as contrastive losses. 

\section{Acknowledgement}
This work was supported by the University of Edinburgh, the Royal Academy of Engineering and Canon Medical Research Europe by a PhD studentship to Xiao Liu. This work was partially supported by the Alan Turing Institute under the EPSRC grant EP/N510129/1. S.A. Tsaftaris acknowledges the support of Canon Medical and the Royal Academy of Engineering and the Research Chairs and Senior Research Fellowships scheme (grant RCSRF1819\textbackslash8\textbackslash25).

\bibliographystyle{splncs04}
\bibliography{references}

\end{document}